\documentclass{article}

\PassOptionsToPackage{numbers, compress}{natbib}
 \usepackage[preprint]{neurips_2026}

\usepackage[utf8]{inputenc} 
\usepackage[T1]{fontenc}    
\usepackage{hyperref}       
\usepackage{url}            
\usepackage{booktabs}       
\usepackage{amsfonts}       
\usepackage{nicefrac}       
\usepackage{microtype}      
\usepackage{xcolor}         
\usepackage{graphicx}
\usepackage{wrapfig}
\usepackage{amsmath}
\usepackage{multirow}
\usepackage[table]{xcolor}
\usepackage{fvextra}
\title{SLVR: Structured Latent Visual Reasoning via Human-like Reasoning Flows}

\author{%
  Bo (Albert) Gao \\
  Carnegie Mellon University \\
  \And
  Bing Xue\thanks{Co-corresponding authors: Bing Xue and Andrea Zanette.} \\
  Meta \\
  \And
  Andrea Zanette\footnotemark[1] \\
  Carnegie Mellon University \\
}

\begin{document}

\maketitle

\begin{abstract}
Multimodal large language models (MLLMs) often answer visual reasoning questions by relying on linguistic priors rather than task-relevant visual evidence. Textual chain-of-thought reasoning can partially mitigate this issue by encouraging models to decompose visual questions into intermediate evidence-seeking steps, but generating these steps autoregressively increases inference cost. Latent reasoning avoids explicit rationale generation, but existing approaches provide limited control over what intermediate states encode, making it difficult to impose separate supervision for planning, grounding, and evidence selection.
We propose Structured Latent Visual Reasoning (SLVR), a training framework that bridges explicit chain-of-thought and latent reasoning by organizing multimodal reasoning into typed latent stages for planning, grounding, evidence selection, and reasoning integration.
 SLVR first trains the model to rely on the image by masking answer-revealing text and contrasting the correct answer with visually plausible distractors. It then organizes reasoning into latent stages for planning, grounding, evidence selection, and integration, supervising each stage with the corresponding signal: plans, boxes, visual evidence, and final rationales. This gives latent reasoning an explicit functional structure while avoiding generated textual chains at inference time.
 Built on Qwen2.5-VL-7B, SLVR improves consistently across multimodal reasoning benchmarks, with absolute gains of +9.4 on MMVP and +14.2 on BLINK Relation, as well as improvements on V*, MathVista, and ChartQA. These results suggest that structured latent supervision can improve fine-grained visual reasoning without the decoding overhead of textual CoT. Project page is available \href{https://bogao-code.github.io/SLVR/}{here}.
\end{abstract}

\section{Introduction}
\label{introduction}

Multimodal large language models (MLLMs), which generate language responses from textual and visual inputs, have become a leading approach to vision-language reasoning. Recent systems such as GPT-4o~\cite{openai2024gpt4o}, Gemini~\cite{gemini2024gemini15}, Qwen-VL~\cite{bai2025qwen25vl}, and GLM-V~\cite{hong2025glm45v} show strong performance across visual question answering and multimodal understanding tasks. However, robust visual reasoning remains difficult~\cite{feng2025visually,mmvp,blink,mmmu}: MLLMs can produce fluent but incorrect answers by relying on linguistic priors or shortcut correlations rather than task-relevant visual evidence. This failure is especially pronounced in fine-grained tasks that require localizing relevant regions, extracting precise evidence, and integrating it with the query.

A common strategy to improve reasoning is to externalize intermediate steps. Chain-of-thought (CoT) reasoning decomposes problems into natural-language reasoning steps~\cite{cot}, and multimodal extensions further represent intermediate reasoning as textual rationales, visual traces, tool calls, or programs~\cite{multimodal_cot,ccot,ddcot,dcot,visual_cot,sketchpad,mint,thinkmorph,deepeyes,codevision}. In training-based variants, these intermediate structures may be supervised with annotated rationales, region boxes, program traces, or tool-execution trajectories, encouraging the model to follow an explicit reasoning path rather than predict the answer directly.
However, explicit reasoning introduces two key limitations. First, it imposes a \emph{representational bottleneck}: high-dimensional multimodal representations must be serialized into discrete text, resulting in lossy compression of fine-grained visual information. Second, it incurs substantial \emph{computational overhead}, as each reasoning step must be generated autoregressively. As a result, explicit CoT can be both inefficient and insufficient for capturing detailed visual relationships.

Latent reasoning~\cite{coconut,soft_thinking,pause_tokens,quietstar,spot,latent_reasoning_survey} addresses these issues by performing intermediate computation directly in continuous hidden states. Recent multimodal approaches extend this paradigm through continuous multimodal representations, visual token manipulation, or latent alignment techniques~\cite{mcout,lvr,ivtlr,lacot,covt,latent_sketchpad,monet,malt,reasoning_mind,tte,hive,lavit}. While these methods improve efficiency and preserve richer representations, they typically treat latent reasoning as \emph{unstructured computation}, without explicitly modeling different functional stages such as planning, grounding, and evidence selection.

In this work, we investigate whether latent multimodal reasoning can benefit from introducing \emph{explicit structure and supervision within the latent space}. We propose \emph{Structured Latent Visual Reasoning (SLVR)}, a training framework that organizes latent reasoning into a sequence of functionally distinct stages and supervises each stage with task-specific signals.

SLVR consists of two complementary components. First, we introduce \emph{Visual Dependency Induction (VDI)}, a contrastive training scheme that reduces reliance on textual shortcuts by removing answer-revealing cues and introducing image-grounded distractors. This encourages the model to rely more on visual evidence during reasoning. Second, we structure the model's internal computation using \emph{typed latent state segments}, implemented via special delimiter tokens (e.g., <plan>, <box>, <evidence>, <reason>, <answer>). These segments partition the latent space into functional roles corresponding to planning, visual grounding, evidence selection, and reasoning integration. During training, each segment is supervised with task-specific targets, including action plan embeddings, bounding box coordinates, and visual patch features. This design enforces a stage-wise decomposition of reasoning while retaining the efficiency of latent computation.

\begin{figure}
    \centering
    \vspace{-1em}
    \includegraphics[width=0.98\linewidth]{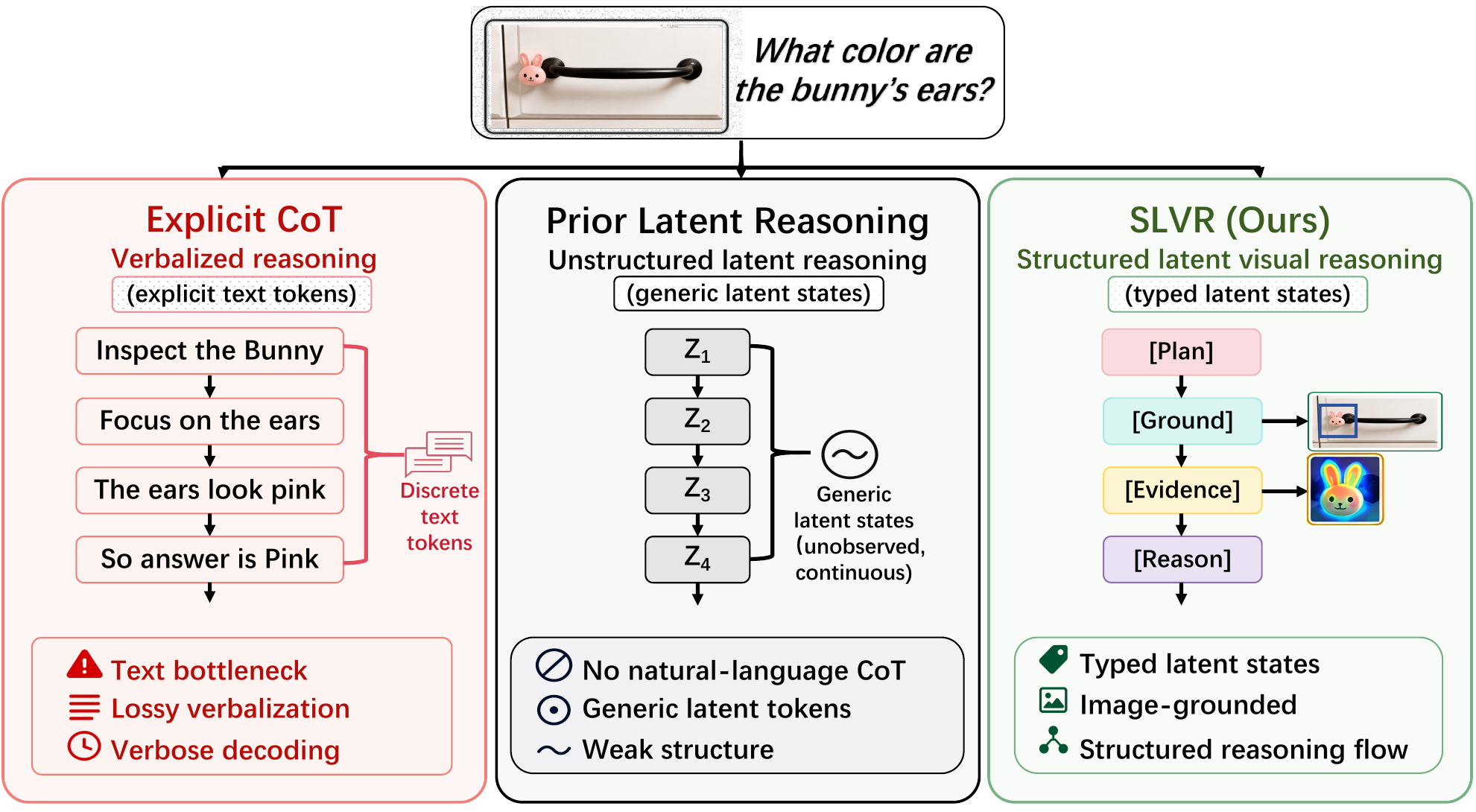}
    \vspace{-0.8em}
    \caption{
\textbf{Comparison of CoT, prior latent reasoning, and SLVR.}
Explicit CoT serializes intermediate reasoning into natural-language tokens, introducing a representational bottleneck and decoding overhead. Prior latent reasoning operates in continuous space but typically treats intermediate states as unstructured. SLVR instead organizes latent computation into typed state segments with distinct functional roles for planning, grounding, evidence selection, and reasoning integration.
}
\vspace{-2.2em}
    \label{fig:teaser}
\end{figure}

We evaluate SLVR on multiple multimodal reasoning benchmarks, including V*, MMVP, BLINK, MathVista, and ChartQA. SLVR consistently improves performance over strong baselines, with particularly large gains on tasks requiring fine-grained spatial and relational reasoning. Notably, when built on Qwen2.5-VL-7B, SLVR achieves improvements of +9.4 on MMVP and +14.2 on BLINK Relation, while avoiding the decoding overhead associated with textual CoT.

Our contributions are summarized as follows:
\begin{itemize}
\item \textbf{A structured alternative to explicit multimodal CoT.}
We introduce a latent reasoning framework that organizes multimodal inference into typed continuous states, providing a compact alternative to decoding long natural-language rationales, visual traces, or program-like reasoning steps.

\item \textbf{Functionally supervised latent reasoning states.}
Rather than treating latent reasoning as an undifferentiated hidden bottleneck, we assign distinct roles to different latent states and supervise them with corresponding targets for planning, grounding, evidence selection, and final answer prediction.

\item \textbf{A shortcut-resistant visual dependency objective.}
We propose visual dependency induction, a contrastive training scheme that removes answer shortcuts from text and uses image-grounded distractors to force reliance on visual evidence.

\item \textbf{Empirical evidence for grounded latent reasoning.}
Experiments across multiple benchmarks show that our framework improves fine-grained visual grounding and compositional reasoning, while reducing reliance on externally decoded reasoning traces.
\end{itemize}

\section{Related Work}
\label{related work}
\subsection{Explicit Multimodal Reasoning}
Recent work extends chain-of-thought~\cite{cot} reasoning to multimodal settings by making intermediate reasoning steps explicit~\cite{mcot_survey}. 
Early multimodal CoT~\cite{multimodal_cot,ccot,ddcot,dcot} methods elicit or supervise textual rationales for visual question answering and compositional reasoning, while later methods expand explicit reasoning into richer multimodal traces, such as visual sketchpads, generated visual thoughts, and interleaved vision-text reasoning steps~\cite{visual_cot,sketchpad,mint,thinkmorph,zhou2024image,chern2025thinking}. 
Another line externalizes visual reasoning through tools or executable programs: methods such as DeepEyes and DeepEyes-v2~\cite{deepeyes,deepeyesv2} actively inspect images with visual tools, while programming-based approaches~\cite{codevision,visual_programming,vipergpt} use code as an interface for cropping, measuring, or analyzing visual content.

These approaches improve interpretability and provide strong supervision by exposing how a model decomposes a visual reasoning problem. 
However, they still depend on externalized intermediate steps, including textual rationales, tool calls, programs, or generated visual traces, which introduce additional decoding overhead and may force early commitment to discrete reasoning paths. 
In contrast, we study whether multimodal reasoning can be organized in latent space, keeping intermediate computation continuous and internal while still grounding it in task-relevant visual evidence.

\subsection{Latent Multimodal Reasoning}
Latent reasoning methods question whether intermediate reasoning must be fully externalized as discrete natural-language tokens. 
Prior work explores continuous hidden-state thoughts~\cite{coconut}, soft concept tokens~\cite{soft_thinking}, pause/thought tokens~\cite{pause_tokens}, and self-generated internal rationales~\cite{quietstar,spot} as alternatives to explicit CoT. 
These studies motivate reasoning as continuous internal computation, a direction also summarized in recent latent reasoning surveys~\cite{latent_reasoning_survey}.

In multimodal settings, this direction is especially appealing because MLLMs already align visual and textual information in a shared latent space. 
Multimodal Coconut~\cite{mcout} extends continuous-thought reasoning to vision-language models. 
Latent Visual Reasoning~\cite{lvr} supervises hidden states using query-relevant ROI visual tokens, while Interleaved Vision-Text Latent Reasoning~\cite{ivtlr} injects selected image tokens into latent reasoning steps. 
LaCoT~\cite{lacot} instead formulates visual reasoning chains as latent variables learned through amortized inference. 
Other methods explore complementary forms of continuous visual thoughts, including compact visual tokens~\cite{covt}, latent visual scratchpads~\cite{latent_sketchpad}, generated visual embeddings~\cite{monet}, modality-agnostic latent tokens~\cite{malt}, and test-time latent think-token optimization~\cite{reasoning_mind}. 
Recent surveys further summarize the broader shift from explicit multimodal CoT toward image-grounded intermediate computation~\cite{continuous_visual_thought_review}.

Our work follows this latent multimodal reasoning direction, but addresses a different limitation: how to organize latent reasoning rather than only how to represent it. 
Prior methods improve latent multimodal computation, but their intermediate latent states are often homogeneous or weakly structured. 
In contrast, we decompose latent visual reasoning into typed functional states for planning, grounding, evidence selection, reasoning integration, and answer decoding. 
This gives the model an explicit inductive bias for task-relevant visual reasoning while preserving the efficiency and flexibility of continuous latent computation. 
A detailed qualitative comparison is provided in Appendix~\ref{app:latent_comparison}.

\section{Method}
\label{method}
\begin{figure*}[t]
    \centering
    \includegraphics[width=0.98\textwidth]{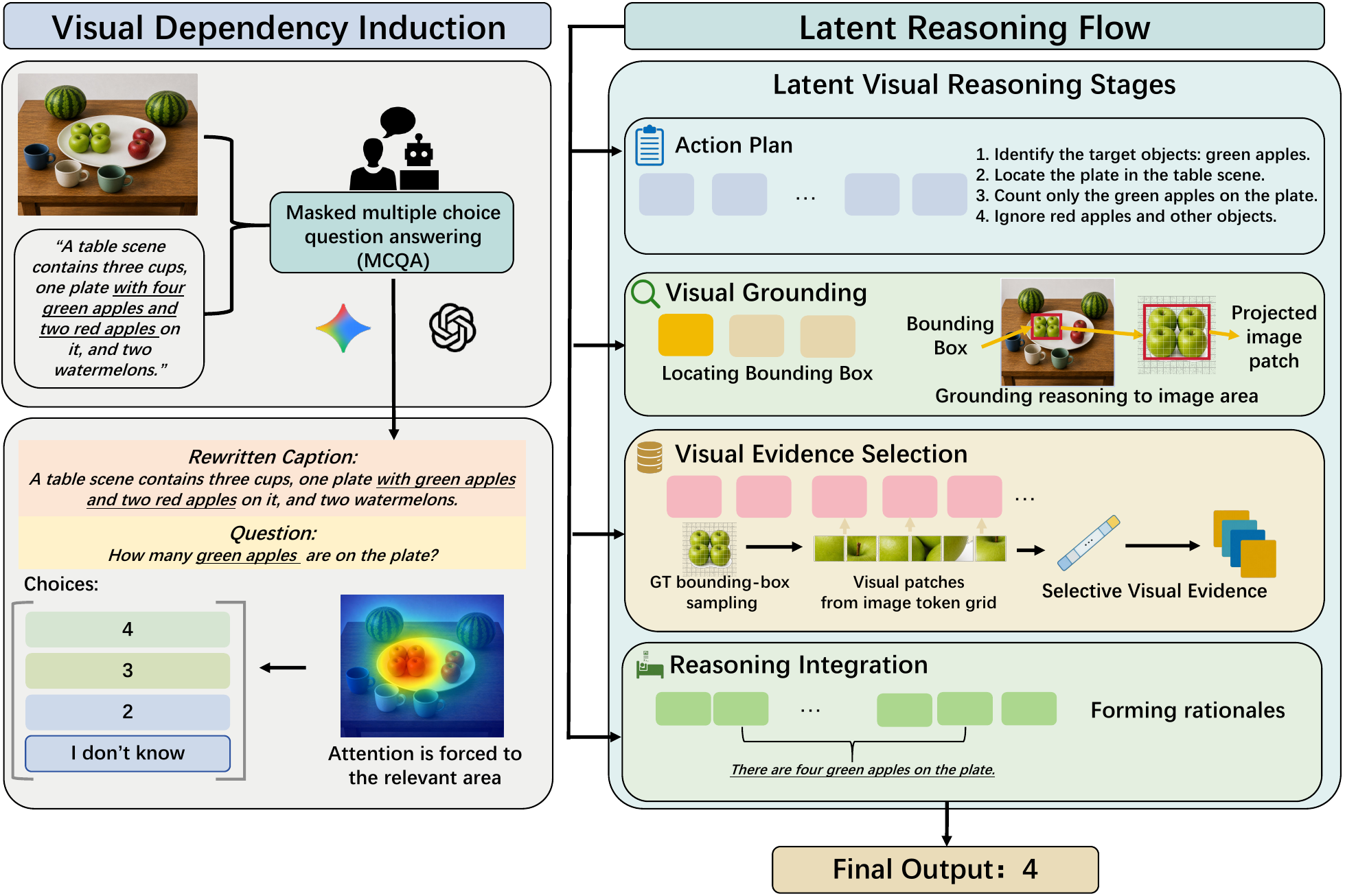}
    \vspace{-0.5em}
    \caption{
    \textbf{Overview of SLVR.}
    SLVR consists of two stages. First, \textit{Visual Dependency Induction} removes answer-revealing textual shortcuts from the caption and constructs masked multiple-choice question answering (MCQA) examples with plausible distractors, encouraging the model to rely on image evidence. Second, the \textit{Latent Reasoning Pipeline} organizes multimodal reasoning into typed latent slots for planning, grounding, visual evidence selection, and reasoning integration. Each stage receives task-specific supervision, including CoT embeddings, bounding boxes, sampled image-token patches, and right answer embeddings.
    }
    \label{fig:pipeline}
    \vspace{-1.5em}
\end{figure*}
\subsection{Overview of SLVR}
We propose \textbf{Structured Latent Visual Reasoning (SLVR)}, a two-stage framework for grounded latent multimodal reasoning, as illustrated in Fig.~\ref{fig:pipeline}. 
The first stage, \textit{Visual Dependency Induction} (Sec.~\ref{stage1}), activates image-grounded reasoning by masking answer-revealing caption cues and introducing plausible image-grounded distractors, forcing the model to resolve the question from visual evidence rather than textual shortcuts and recover the missing fact from the image.  
The second stage, \textit{Structured Latent Visual Reasoning} (Sec.~\ref{stage2}), organizes the model's hidden states into a typed latent reasoning flow before answer generation, with functional latent stages for planning, grounding, visual evidence selection and reasoning integration. 
Together, these two stages first establish visual dependency and then structure it in latent space, preserving the compactness of latent reasoning while avoiding homogeneous, unstructured latent tokens. Their effectiveness is evaluated in Sec 4.4. 

\subsection{Intuition and Example} 
Consider the scenario illustrated in Figure~\ref{fig:pipeline}: the original caption states that the scene contains ``one plate with four green apples and two red apples,'' which directly reveals the answer to the question ``How many green apples are on the plate?'' VDI removes this answer-revealing cue, rewriting the caption as a partial description (e.g., ``one plate with green and red apples''). The question remains unchanged, but the answer options are populated with hard, image-grounded distractors. The correct answer is \emph{4}, while distractors such as \emph{2} and \emph{3} correspond to other visible quantities in the scene (e.g., red apples or cups). Unlike random negatives, these distractors are visually plausible yet incorrect under the query, encouraging the model to rely on visual evidence rather than textual shortcuts.

Building on this grounding, SLVR further structures how the model performs reasoning internally. Given the same example, the model processes the input through a sequence of latent stages. In the \texttt{<plan>} stage, it encodes a high-level strategy (e.g., counting green apples on the plate). In the \texttt{<box>} stage, it predicts a spatial region corresponding to the plate. In the \texttt{<evidence>} stage, it aggregates visual features from image patches within this region that correspond to green apples. Finally, in the \texttt{<reason>} and \texttt{<answer>} stages, the model integrates the extracted evidence to produce the final count. Each stage operates in continuous latent space and is supervised with task-specific targets, enabling structured multi-step reasoning without requiring explicit textual intermediate steps.

\subsection{Visual Dependency Induction}
\label{stage1}
\textbf{Formalization.}
We first introduce \textit{Visual Dependency Induction} (VDI), an initial MCQA training stage that converts coarse image-caption alignment into question-conditioned visual dependency.
Each raw instance is an image-caption pair $(I,c)$, where $c$ is broadly aligned with the visual content.
We use a caption-rewriting procedure to select an answer-critical visual fact $e$ from $c$ and produce a redacted context $c^{-}$ by removing only $e$ while preserving the remaining grounded description.
A query $q$ and ground-truth answer $y$ are then generated from the removed fact $e$.
We construct a candidate set $\mathcal{A}=\{y\}\cup\mathcal{D}$, where $\mathcal{D}$ contains image-grounded distractors derived from other objects, attributes, counts, or relations present within $I$.

For each candidate $a\in\mathcal{A}$, the model receives the multimodal prompt tuple $(I, c^{-}, q)$ as input and generates the candidate text $a$. 
We define the compatibility score $s_\theta$ as the log-likelihood of generating candidate $a$ given the prompt:
\begin{equation}
s_\theta(I,c^{-},q,a)
=
\log p_\theta(a \mid I,c^{-},q).
\end{equation}
We then train the network using a softmax contrastive objective over these compatibility scores:
\begin{equation}
\mathcal{L}_{\mathrm{VDI}}
=
-\log
\frac{
\exp s_\theta(I,c^{-},q,y)
}{
\sum_{a \in \mathcal{A}}
\exp s_\theta(I,c^{-},q,a)
}.
\label{eq:vdi_loss}
\end{equation}
By explicitly contrasting the correct answer against visually grounded but question-inconsistent alternatives, VDI compels the model to form representations that are strictly dependent on task-relevant visual evidence. This induction establishes the critical grounding foundation required for the subsequent structured latent reasoning stage.

\subsection{Structured Latent Visual Reasoning}
\label{stage2}
\paragraph{Latent reasoning flow.}
With visual dependency established, SLVR organizes the model's internal computation into a structured latent reasoning trajectory. Rather than compressing a visual chain-of-thought into a monolithic latent vector or an unstructured token sequence, we enforce a human-like cognitive path natively within the latent space. To make this functional structure concrete, we format each training instance by inserting an ordered sequence of typed latent states and transition delimiters prior to the final answer:

\begin{equation}
\small
\begin{split}
S = [&
\langle \mathrm{plan\_start} \rangle \; Z_{\mathrm{plan}} \; \langle \mathrm{plan\_end} \rangle,\;
\langle \mathrm{box\_start} \rangle \; Z_{\mathrm{box}} \; \langle \mathrm{box\_end} \rangle, \\
&
\langle \mathrm{evi\_start} \rangle \; Z_{\mathrm{evi}} \; \langle \mathrm{evi\_end} \rangle,\;
\langle \mathrm{full\_start} \rangle \; Z_{\mathrm{full}} \; \langle \mathrm{full\_end} \rangle,\;
\langle \mathrm{ans} \rangle \; y
].
\end{split}
\label{eq:latent_sequence}
\end{equation}
The special delimiter tokens are not merely formatting markers. They define the boundaries and types of latent states, and they are trained with next-token prediction so that the model learns an ordered transition structure: planning $\rightarrow$ grounding $\rightarrow$ evidence selection $\rightarrow$ reasoning integration $\rightarrow$ answer decoding. In contrast, the latent state tokens themselves are masked from the standard language modeling loss and are supervised through state-specific alignment objectives.

\paragraph{State roles.}
The latent states in Eq.~\ref{eq:latent_sequence} serve as internal reasoning states with distinct semantic roles. The planning state $Z_{\mathrm{plan}}$ represents \emph{what to reason about}, such as whether the question requires counting, attribute recognition, spatial comparison, or object presence verification. The grounding state $Z_{\mathrm{box}}$ represents \emph{where to look} by predicting a target region in normalized bounding-box coordinates. The evidence state $Z_{\mathrm{evi}}$ represents \emph{what visual evidence to use} by carrying answer-relevant visual patch information from the grounded region. Finally, the integration state $Z_{\mathrm{full}}$ represents \emph{how to integrate the reasoning process}, serving as a compact latent substitute for the full visual rationale before the model produces the short answer $y$.

This design separates two complementary learning signals: delimiter tokens teach the model the discrete reasoning-state transition, while latent state tokens learn the continuous content of each reasoning state. As a result, SLVR does not rely on a homogeneous block of latent computation, but instead induces a structured latent trajectory with explicit stage boundaries.

\textbf{State-Specific Latent Alignment.} We optimize the continuous latent states through a unified, state-specific alignment objective. The core intuition is to constrain each latent state to approximate a supervision target that strictly corresponds to its functional role in the reasoning trajectory. For example, the planning state is aligned with the textual embedding of an action plan, the grounding state with spatial bounding-box coordinates, the evidence state with relevant visual patch features, and the integration state with the embedding of the complete visual rationale.

Formally, let $m$ denote a latent state type, with
\[
\mathcal{M}=\{\mathrm{plan}, \mathrm{box}, \mathrm{evi}, \mathrm{full}\}.
\]
For each type $m$, $H_{m}$ denotes the hidden states at the corresponding latent positions, $P_{m}$ is a type-specific projection head, and $T_{m}$ is the supervision target for that type. We define the latent alignment loss as:
\begin{equation}
\begin{aligned}
\mathcal{L}_{\mathrm{SLA}}
&=
\sum_{m \in \mathcal{M}}
\lambda_{m}
D_{m}
\left(
P_{m}(H_{m}), T_{m}
\right) \\
&=
\lambda_{\mathrm{plan}}
\left\|
P_{\mathrm{plan}}(H_{\mathrm{plan}})-T_{\mathrm{plan}}
\right\|_2^2
+
\lambda_{\mathrm{box}}
\left\|
P_{\mathrm{box}}(H_{\mathrm{box}})-T_{\mathrm{box}}
\right\|_2^2 \\
&\quad+
\lambda_{\mathrm{evi}}
\left\|
P_{\mathrm{evi}}(H_{\mathrm{evi}})-T_{\mathrm{evi}}
\right\|_2^2
+
\lambda_{\mathrm{full}}
\left\|
P_{\mathrm{full}}(H_{\mathrm{full}})-T_{\mathrm{full}}
\right\|_2^2 .
\end{aligned}
\label{eq:sla_loss}
\end{equation}

While all terms optimize a standard $\ell_2$ distance, they operate across distinct representational spaces to enforce functional modularity. 
The \textbf{planning} latent tokens are aligned with the embedding of the textual action plan, encouraging them to encode \emph{what to reason about}. 
The \textbf{grounding} latent tokens are aligned with the normalized bounding box $b=(c_x,c_y,w,h)$, encouraging them to encode \emph{where to look}. Specifically, we allocate three latent tokens for this state: the first predicts the object center $(c_x,c_y)$, while the second and third predict the boundary (width $w$ and height $h$, respectively). 
The \textbf{evidence} latent tokens are aligned with selected, query-relevant visual patch features from the grounded region, encouraging them to encode \emph{which visual evidence to use}. 
Finally, the \textbf{integration} latent tokens are aligned with the embedding of the complete visual rationale, encouraging them to synthesize \emph{how to logically deduce} the final answer.

\textbf{Evidence Target Generation.} The evidence state $Z_{\mathrm{evi}}$ is designed to refine coarse spatial grounding into precise, answer-relevant visual features. However, because a ground-truth bounding box may still encompass background patches, irrelevant object parts, or distractor details, we must construct a refined supervision target for $Z_{\mathrm{evi}}$ during training.

We achieve this by utilizing the final answer to heuristically select the most relevant visual patches. Let $\mathcal{B}$ denote the set of visual patch indices overlapping with the ground-truth box, and let $v_i \in \mathbb{R}^{d}$ be the visual feature of patch $i$. During training, we encode the full answer $y$ into an answer query vector $a \in \mathbb{R}^{d}$. We then compute a dot-product relevance score between each candidate patch and the answer query:
\begin{equation}
s_i = v_i^\top a, \quad i \in \mathcal{B}.
\label{eq:evidence_score}
\end{equation}
In practice, we normalize both vectors before scoring, yielding a normalized dot product:
\begin{equation}
s_i =
\left(
\frac{v_i}{\|v_i\|_2}
\right)^\top
\left(
\frac{a}{\|a\|_2}
\right),
\quad i \in \mathcal{B}.
\label{eq:evidence_score_norm}
\end{equation}

We select patches whose semantic relevance score $s_i$ exceeds a predefined threshold $\tau_{\mathrm{evi}}$ to form the evidence supervision target. To prevent empty or overly sparse evidence sets, we enforce a minimum patch count $K_{\min}$ by falling back to the top-ranked candidates if necessary:
\begin{equation}
\mathcal{P}_{\mathrm{evi}}
=
\begin{cases}
\{\, i \in \mathcal{B} \mid s_i \geq \tau_{\mathrm{evi}} \,\}, 
& \text{if } 
\left|\{\, i \in \mathcal{B} \mid s_i \geq \tau_{\mathrm{evi}} \,\}\right| \geq K_{\min}, \\[3pt]
\operatorname{TopK}_{i \in \mathcal{B}}(s_i, K_{\min}),
& \text{otherwise}.
\end{cases}
\end{equation}
\begin{equation}
T_{\mathrm{evi}}
=
\{\, v_i \mid i \in \mathcal{P}_{\mathrm{evi}} \,\}.
\label{eq:evidence_target}
\end{equation}

It is crucial to emphasize that this dot-product selection mechanism is strictly a \textbf{training-time procedure} used to formulate the supervision target $T_{\mathrm{evi}}$. At inference time, the model does not have access to the answer query $a$. Instead, having been optimized via the alignment loss $\mathcal{L}_{\mathrm{SLA}}$, the network autonomously projects the latent state $Z_{\mathrm{evi}}$ to implicitly isolate and attend to the correct query-relevant image patches.

\textbf{Training Objective.}
The total training objective combines standard autoregressive supervision,
state-specific latent alignment, and answer-transition supervision:
\begin{equation}
\mathcal{L}
=
\mathcal{L}_{\mathrm{ce}}
+
\mathcal{L}_{\mathrm{SLA}}
+
\mathcal{L}_{\mathrm{ans}} .
\label{eq:total_loss}
\end{equation}
Here, $\mathcal{L}_{\mathrm{SLA}}$ is the state-specific latent alignment objective defined in Eq.~\ref{eq:sla_loss}, which supervises the planning, box-grounding, evidence, and full-answer latent states.

The answer-side objective is defined as:
\begin{equation}
\mathcal{L}_{\mathrm{ans}}
=
\lambda_{\mathrm{open}}\mathcal{L}_{\mathrm{answer\_open}}
+
\lambda_{\mathrm{short}}\mathcal{L}_{\mathrm{short\_answer}}.
\end{equation}
$\mathcal{L}_{\mathrm{answer\_open}}$ teaches the model when to stop internal latent reasoning and transition to the explicit \texttt{<answer>} segment, while $\mathcal{L}_{\mathrm{short\_answer}}$ supervises the concise final answer produced after this transition.

During training, $\mathcal{L}_{\mathrm{ce}}$ is applied to the discrete transition delimiters and output tokens, while the continuous latent states are optimized through $\mathcal{L}_{\mathrm{SLA}}$ rather than standard next-token prediction.

At inference time, the model operates autonomously without access to any auxiliary targets. Given only the input image and question, the network autoregressively decodes the discrete transition delimiters while internally projecting the corresponding continuous latent states to execute planning, grounding, evidence extraction, and integration. Following this structured internal rollout, the model decodes a concise final answer. Thus, SLVR exploits rich, structured multimodal supervision during training while maintaining a highly efficient, direct-answer interface during deployment.

\begin{table*}[t]
\centering
\small
\setlength{\tabcolsep}{4.8pt}
\renewcommand{\arraystretch}{1.08}
\caption{
\textbf{Performance on challenging multimodal perception and reasoning benchmarks.}
\textbf{V*} and \textbf{BLINK} are reported with their fine-grained subsets, while \textbf{MMVP} is reported using the standard overall accuracy.
The best result for each metric is highlighted in \textbf{bold}.
The green row reports the absolute improvement of \textbf{SLVR} over the \textbf{Qwen2.5-VL-7B} base model. Prior-method results are taken from the corresponding papers. 
}
\label{tab:main_results}
\resizebox{\textwidth}{!}{
\begin{tabular}{lcccccccccc}
\toprule
\multirow{2}{*}{\textbf{Model}}
& \multicolumn{3}{c}{\textbf{V*}}
& \multirow{2}{*}{\textbf{MMVP}}
& \multicolumn{4}{c}{\textbf{BLINK}}
& \multirow{2}{*}{\textbf{MathVista}}
& \multirow{2}{*}{\textbf{ChartQA}} \\
\cmidrule(lr){2-4}
\cmidrule(lr){6-9}
& \textbf{Overall} & \textbf{Attribute} & \textbf{Spatial}
&
& \textbf{Count} & \textbf{Jigsaw} & \textbf{Relation} & \textbf{IQ}
&
\\
\midrule

\textbf{Qwen2.5-VL-7B}
& 78.5 & 81.7 & 73.7
& 66.7
& 66.7 & 52.0 & 38.8 & 26.0
& 63.7 & 86.2 \\

\textbf{Monet}$^\dagger$~\cite{monet}
& 83.6 & 83.5 & 82.9
& --
& -- & -- & -- & --
& -- & -- \\

\textbf{LVR}$^\dagger$~\cite{lvr}
& 80.6 & 81.7 & 79.0
& 71.8
& 70.0 & 52.5 & 42.3 & 28.4
& -- & -- \\

\textbf{DeepEyes v2}$^\dagger$~\cite{deepeyesv2}
& 81.8 & -- & --
& --
& -- & -- & -- & --
& 71.9 & 88.4 \\

\textbf{COVT}$^\dagger$~\cite{covt}
& 78.5 & -- & --
& 56.7
& -- & -- & -- & --
& -- & -- \\

\textbf{CodeVision}$^\dagger$~\cite{codevision}
& 83.7 & -- & --
& --
& -- & -- & -- & --
& -- & -- \\

\textbf{LaViT}$^\dagger$~\cite{lavit}
& -- & 82.6 & --
& 67.3
& -- & -- & 45.5 & 32.0
& -- & -- \\

\midrule
\rowcolor{green!8}
\textbf{SLVR (Ours)}
& \textbf{85.2}
& \textbf{85.1}
& \textbf{84.3}
& \textbf{76.1}
& \textbf{72.4}
& \textbf{62.3}
& \textbf{53.0}
& \textbf{34.5}
& \textbf{73.4}
& \textbf{89.6} \\

\rowcolor{green!8}
\textit{Improvement over Qwen}
& \textcolor{green!50!black}{+6.7}
& \textcolor{green!50!black}{+3.4}
& \textcolor{green!50!black}{+10.6}
& \textcolor{green!50!black}{+9.4}
& \textcolor{green!50!black}{+5.7}
& \textcolor{green!50!black}{+10.3}
& \textcolor{green!50!black}{+14.2}
& \textcolor{green!50!black}{+8.5}
& \textcolor{green!50!black}{+9.7}
& \textcolor{green!50!black}{+3.4} \\

\bottomrule
\end{tabular}
}
\vspace{-2em}
\end{table*}

\begin{table*}[t]
\centering
\small
\setlength{\tabcolsep}{4.0pt}
\renewcommand{\arraystretch}{1.08}
\caption{
\textbf{Ablation studies of SLVR.}
Left: incremental ablation of the training pipeline on \textbf{MMVP} and \textbf{V*}.
Right: ablation of latent token budgets, where \textbf{P}, \textbf{E}, and \textbf{I} denote the number of latent tokens for planning, evidence selection, and reasoning integration, respectively.
}
\label{tab:ablation}
\resizebox{\textwidth}{!}{
\begin{tabular}{lcccc|lccccc}
\toprule
\multicolumn{5}{c|}{\textbf{Incremental Components}}
& \multicolumn{6}{c}{\textbf{Latent Token Budget}} \\
\midrule
\multirow{2}{*}{\textbf{Variant}}
& \multirow{2}{*}{\textbf{MMVP}}
& \multicolumn{3}{c|}{\textbf{V*}}
& \multirow{2}{*}{\textbf{Setting}}
& \multirow{2}{*}{\textbf{P}}
& \multirow{2}{*}{\textbf{E}}
& \multirow{2}{*}{\textbf{I}}
& \multirow{2}{*}{\textbf{MMVP}}
& \multirow{2}{*}{\textbf{V* Overall}} \\
\cmidrule(lr){3-5}
& & \textbf{Overall} & \textbf{Attr.} & \textbf{Spatial}
& & & & & & \\
\midrule

\textbf{Qwen2.5-VL-7B}
& 66.7 & 78.5 & 81.7 & 73.7
& \textbf{Short latent}
& 16 & 16 & 16 & 74.0 & 83.8 \\

\textbf{+ Masked Text}
& 71.2 & 81.9 & 83.0 & 80.1
& \textbf{Long grounding}
& 64 & 64 & 16 & 74.8 & 84.3 \\

\textbf{+ Latent Plan}
& 73.0 & 83.0 & 83.7 & 81.5
& \textbf{Long integration}
& 64 & 16 & 64 & 74.5 & 84.1 \\

\textbf{+ Latent Grounding}
& 74.2 & 84.0 & 84.3 & 82.9
& \textbf{Balanced}
& 32 & 32 & 32 & 75.1 & 84.5 \\

\textbf{+ Latent Evidence}
& 75.2 & 84.6 & 84.8 & 83.6
& \cellcolor{green!8}\textbf{Ours}
& \cellcolor{green!8}\textbf{64}
& \cellcolor{green!8}\textbf{16}
& \cellcolor{green!8}\textbf{16}
& \cellcolor{green!8}\textbf{76.1}
& \cellcolor{green!8}\textbf{85.2} \\

\rowcolor{green!8}
\textbf{+ Latent Integration / SLVR}
& \textbf{76.1} & \textbf{85.2} & \textbf{85.1} & \textbf{84.3}
& \textit{Improvement over Qwen}
& -- & -- & --
& \textcolor{green!50!black}{+9.4}
& \textcolor{green!50!black}{+6.7} \\

\bottomrule
\end{tabular}
}
\vspace{-2em}
\end{table*}

\section{Experiments}
\label{Experiments}
\subsection{Experimental Setup}
\textbf{Backbone model.}
We use \textbf{Qwen2.5-VL-7B-Instruct} as the backbone model for SLVR.
This backbone has been adopted by several recent open-source multimodal reasoning methods~\cite{monet,lvr}, which makes it a natural choice for controlled comparison.
To ensure fairness, we keep the backbone architecture and initialization fixed across our experiments.
Therefore, performance differences are less likely to arise from different model scales or pretrained visual encoders, and can be more directly attributed to our proposed visual dependency induction and structured latent reasoning training. 

\textbf{Training data.}
SLVR is trained with a two-stage data pipeline.
For the first stage, we construct a \textbf{20K masked-context dataset} from COVT~\cite{covt} visual reasoning data.
Each raw instance is an aligned image-text pair $(I,c)$, where $c$ describes concrete visual facts such as objects, attributes, counts, or relations.
We use an off-the-shelf VLM-assisted rewriting procedure to identify an answer-critical visual fact in $c$, remove only this fact while preserving the remaining grounded context, and generate the corresponding question, answer, and multiple-choice options.
The resulting redacted context remains coherent but no longer reveals the answer, forcing the model to recover the missing fact from the image.
Distractor options are visually plausible but incorrect under the generated query.
This procedure is used only for training-data construction, not at inference time, and does not provide teacher predictions for evaluation examples.
We manually verify a subset of the constructed examples and filter cases with ambiguous evidence, incorrect masking, answer leakage, or non-unique answers.
More details are provided in Appendix~\ref{app:vdi}.

For the second stage, we train on Visual-CoT~\cite{visual_cot} to learn structured latent visual reasoning.
Visual-CoT provides image-question-answer examples with CoT-style plans, region-level grounding annotations, and full reasoning/answer supervision.
We use these annotations to supervise the typed latent states for \textit{planning}, \textit{grounding}, \textit{evidence selection}, and \textit{reasoning integration}, while avoiding the need to decode full visible CoT at inference time.
More details about the Visual-CoT data format, latent-state construction, and training targets are provided in Appendix~\ref{app:visual_cot_format}.

\textbf{Evaluation benchmarks.}
We evaluate SLVR on five challenging multimodal perception and reasoning benchmarks: \textbf{V*}~\cite{v_star}, \textbf{MMVP}~\cite{mmvp}, \textbf{BLINK}~\cite{blink}, \textbf{MathVista}~\cite{math}, and \textbf{ChartQA}~\cite{chartqa}.
For \textbf{V*}, we report the overall score as well as the attribute and spatial subsets.
For \textbf{BLINK}~\cite{blink}, we report fine-grained subsets including counting, jigsaw, relation, and IQ-style visual reasoning.
For \textbf{MMVP}~\cite{mmvp}, we report the standard overall accuracy.
Together, these benchmarks evaluate fine-grained visual perception, spatial reasoning, visual robustness, and multimodal reasoning beyond language-only shortcuts.

\subsection{Implementation Details}
\label{Implementation Details}
We initialize SLVR from Qwen2.5-VL-7B-Instruct and train it with a two-stage procedure.
All experiments are conducted on 8 NVIDIA A100 GPUs with mixed-precision (bf16) training.
During training, we freeze the visual encoder and image processor, and update only the language model parameters.
No additional vision encoder or projection module is trained.
Images are processed with the default Qwen2.5-VL image processor.

We use a learning rate of $1\times10^{-5}$ for the first-stage masked-text training and $5\times10^{-6}$ for the second-stage latent reasoning training.
The global batch size is set to 64 with gradient accumulation.
We train the first stage for 5 epochs and the second stage for 3 epochs.
The maximum sequence length is set to 4096.
We use AdamW as the optimizer with a cosine learning-rate schedule and a warmup ratio of 0.03.
The weight decay is set to 0.0.

In the inference stage, we set the number of latent tokens for planning, evidence modeling, and answer integration to 64, 16, and 16, respectively. Grounding uses a fixed 3 tokens as defined in Sec ~\ref{stage2}.
For evidence patch selection, we use an evidence threshold of $\tau_{\mathrm{evi}}=0.5$ and keep at least $K_{\min}=5$ patches by falling back to the top-ranked candidates when necessary.
For the overall training objective, we use fixed empirical weights:
\begin{equation}
\mathcal{L}
=
\mathcal{L}_{\mathrm{ce}}
+ 0.1\mathcal{L}_{\mathrm{plan}}
+ 1.0\mathcal{L}_{\mathrm{box}}
+ 1.0\mathcal{L}_{\mathrm{evi}}
+ 0.5\mathcal{L}_{\mathrm{full}}
+ 2.0\mathcal{L}_{\mathrm{answer\_open}}
+ 3.0\mathcal{L}_{\mathrm{short\_answer}}.
\end{equation}
These hyperparameter choices are analyzed in Appendix~\ref{app:hyperparameter_analysis}.

\subsection{Main Results}
\label{Main Results}
Table~\ref{tab:main_results} reports the main comparison on five multimodal perception and reasoning benchmarks.
Compared with the Qwen2.5-VL-7B base model, SLVR improves consistently across all reported metrics.
On \textbf{V*}~\cite{v_star}, SLVR improves Overall, Attribute, and Spatial by 
\textcolor{green!50!black}{+6.7}, 
\textcolor{green!50!black}{+3.4}, and 
\textcolor{green!50!black}{+10.6}, respectively.
On \textbf{MMVP}~\cite{mmvp}, SLVR achieves a substantial gain of 
\textcolor{green!50!black}{+9.4}, suggesting that visual dependency induction helps reduce reliance on language-prior shortcuts and encourages stronger image-grounded reasoning.

SLVR also improves all fine-grained \textbf{BLINK}~\cite{blink} subsets, with gains of 
\textcolor{green!50!black}{+5.7} on Counting, 
\textcolor{green!50!black}{+10.3} on Jigsaw, 
\textcolor{green!50!black}{+14.2} on Relation, and 
\textcolor{green!50!black}{+8.5} on IQ.
The strong improvements on Spatial and Relation-oriented subsets indicate that typed latent grounding and evidence selection are particularly beneficial for fine-grained visual reasoning.
Beyond natural-image perception, SLVR further improves \textbf{MathVista}~\cite{math} and \textbf{ChartQA}~\cite{chartqa} by 
\textcolor{green!50!black}{+9.7} and 
\textcolor{green!50!black}{+3.4}, respectively, showing that the learned latent reasoning structure also transfers to broader multimodal reasoning tasks.

\subsection{Ablation Study}
\label{Ablation Study}
Table~\ref{tab:ablation} presents two ablation analyses.
On the left, we progressively add each component on top of Qwen2.5-VL-7B.
The first stage already brings clear gains, improving MMVP~\cite{mmvp} from 66.7 to 71.2 and V*~\cite{v_star} Overall from 78.5 to 81.9.
Adding the typed latent stages further improves performance step by step, with the full SLVR model reaching 76.1 on MMVP~\cite{mmvp} and 85.2 on V*~\cite{v_star} Overall.
These results show that visual dependency induction and structured latent reasoning are complementary.

On the right of Table 2, we study the effect of latent token budgets. We denote the number of latent tokens for planning, evidence selection, and reasoning integration as $P$, $E$, and $I$, respectively. As shown, SLVR performs optimally with an asymmetric token distribution ($P=64$, $E=16$, $I=16$). This distribution mirrors the informational density of our supervision targets. The planning stage is supervised by natural language CoT rationales, which have a mean length of 65.11 tokens (median 62). Allocating 64 latent tokens provides a nearly 1:1 representational capacity to align with these complex syntactic structures. In contrast, once the global strategy is formulated, the downstream targets---such as low-dimensional spatial coordinates for grounding and highly localized patch features for evidence---require significantly less capacity. Consequently, these later functional states can be aggressively compressed to 16 tokens without performance degradation, maximizing inference efficiency while ensuring both the structure and allocation of latent states are fully optimized.

\textbf{Robustness to misleading textual context.}
We additionally evaluate grounding robustness under misleading captions.
Specifically, we sample non-overlapping COVT~\cite{covt} examples and use a VLM-assisted rewriting procedure to deliberately replace the answer-critical visual fact in the caption with an incorrect but plausible alternative.
As reported in Appendix~\ref{app:misleading_caption}, SLVR achieves higher misleading-context accuracy and lower caption-following error than Qwen2.5-VL-7B, VDI-only, and LVR~\cite{lvr}, suggesting that the proposed structured latent flow better recovers answer-relevant evidence from the image when text and vision conflict.

\section{Conclusion and Limitations}
\label{conclusion and limitation}
We introduce \textbf{SLVR}, a framework that enables MLLMs to perform structured visual reasoning in the latent space.
Our visual dependency induction stage constructs 20K COVT-rewritten masked-text examples to reduce reliance on textual shortcuts, while our structured latent reasoning stage uses Visual-CoT supervision to organize internal reasoning into planning, grounding, evidence selection, and integration.
Experiments show that SLVR consistently improves over Qwen2.5-VL-7B on multimodal benchmarks, with strong gains on fine-grained spatial and relational reasoning.
Together, these results demonstrate SLVR as a promising direction toward more grounded and reliable multimodal reasoning.

We summarize the limitations of our work as follows.
First, SLVR relies on a two-stage training pipeline, which increases data construction and training complexity.
Second, we have not yet explored adaptive latent-state designs or alternative supervision signals for latent visual reasoning.
Third, our experiments are based on Qwen2.5-VL-7B, leaving broader scaling and architecture transfer for future work.

\bibliography{main}

@article{cot,
  title={Chain-of-thought prompting elicits reasoning in large language models},
  author={Wei, Jason and Wang, Xuezhi and Schuurmans, Dale and Bosma, Maarten and Xia, Fei and Chi, Ed and Le, Quoc V and Zhou, Denny and others},
  journal={Advances in neural information processing systems},
  volume={35},
  pages={24824--24837},
  year={2022}
}

@inproceedings{deepeyes,
title={DeepEyes: Incentivizing ''Thinking with Images'' via Reinforcement Learning},
author={Ziwei Zheng and Michael Yang and Jack Hong and Chenxiao Zhao and Guohai Xu and Le Yang and Chao Shen and XingYu},
booktitle={The Fourteenth International Conference on Learning Representations},
year={2026},
}

@inproceedings{ deepeyesv2,
title={DeepEyesV2: Toward Agentic Multimodal Model},
author={Jack Hong and Chenxiao Zhao and Chenglin Zhu and Weiheng Lu and Guohai Xu and XingYu},
booktitle={The Fourteenth International Conference on Learning Representations},
year={2026},
}

@article{codevision,
  title={Thinking with Programming Vision: Towards a Unified View for Thinking with Images},
  author={Guo, Zirun and Hong, Minjie and Zhang, Feng and Jia, Kai and Jin, Tao},
  journal={arXiv preprint arXiv:2512.03746},
  year={2025}
}

@inproceedings{thinkmorph,
title={ThinkMorph: Emergent Properties in Multimodal Interleaved Chain-of-Thought Reasoning},
author={Jiawei Gu and Yunzhuo Hao and Huichen Will Wang and Linjie Li and Michael Qizhe Shieh and Yejin Choi and Ranjay Krishna and Yu Cheng},
booktitle={The Fourteenth International Conference on Learning Representations},
year={2026},
}

@article{mcot_survey,
  title={Multimodal chain-of-thought reasoning: A comprehensive survey},
  author={Wang, Yaoting and Wu, Shengqiong and Zhang, Yuecheng and Yan, Shuicheng and Liu, Ziwei and Luo, Jiebo and Fei, Hao},
  journal={arXiv preprint arXiv:2503.12605},
  year={2025}
}

@article{multimodal_cot,
  title={Multimodal chain-of-thought reasoning in language models},
  author={Zhang, Zhuosheng and Zhang, Aston and Li, Mu and Zhao, Hai and Karypis, George and Smola, Alex},
  journal={arXiv preprint arXiv:2302.00923},
  year={2023}
}

@inproceedings{ccot,
  title={Compositional chain-of-thought prompting for large multimodal models},
  author={Mitra, Chancharik and Huang, Brandon and Darrell, Trevor and Herzig, Roei},
  booktitle={Proceedings of the IEEE/CVF Conference on Computer Vision and Pattern Recognition},
  pages={14420--14431},
  year={2024}
}

@article{ddcot,
  title={Ddcot: Duty-distinct chain-of-thought prompting for multimodal reasoning in language models},
  author={Zheng, Ge and Yang, Bin and Tang, Jiajin and Zhou, Hong-Yu and Yang, Sibei},
  journal={Advances in Neural Information Processing Systems},
  volume={36},
  pages={5168--5191},
  year={2023}
}

@inproceedings{dcot,
  title={Dcot: Dual chain-of-thought prompting for large multimodal models},
  author={Jia, Zixi and Liu, Jiqiang and Li, Hexiao and Liu, Qinghua and Gao, Hongbin},
  booktitle={The 16th Asian Conference on Machine Learning (Conference Track)},
  year={2024}
}

@inproceedings{visual_cot,
title={Visual CoT: Advancing Multi-Modal Language Models with a Comprehensive Dataset and Benchmark for Chain-of-Thought Reasoning},
author={Hao Shao and Shengju Qian and Han Xiao and Guanglu Song and Zhuofan Zong and Letian Wang and Yu Liu and Hongsheng Li},
booktitle={The Thirty-eight Conference on Neural Information Processing Systems Datasets and Benchmarks Track},
year={2024},
}

@inproceedings{sketchpad,
title={Visual Sketchpad: Sketching as a Visual Chain of Thought for Multimodal Language Models},
author={Yushi Hu and Weijia Shi and Xingyu Fu and Dan Roth and Mari Ostendorf and Luke Zettlemoyer and Noah A. Smith and Ranjay Krishna},
booktitle={The Thirty-eighth Annual Conference on Neural Information Processing Systems},
year={2024},
}

@inproceedings{mint,
title={{MINT}-CoT: Enabling Interleaved Visual Tokens in Mathematical Chain-of-Thought Reasoning},
author={Xinyan Chen and Renrui Zhang and Dongzhi Jiang and Aojun Zhou and Shilin Yan and Weifeng Lin and Hongsheng Li},
booktitle={The Thirty-ninth Annual Conference on Neural Information Processing Systems},
year={2026},
}

@article{zhou2024image,
  title={Image-of-thought prompting for visual reasoning refinement in multimodal large language models},
  author={Zhou, Qiji and Zhou, Ruochen and Hu, Zike and Lu, Panzhong and Gao, Siyang and Zhang, Yue},
  journal={arXiv preprint arXiv:2405.13872},
  year={2024}
}

@article{chern2025thinking,
  title={Thinking with generated images},
  author={Chern, Ethan and Hu, Zhulin and Chern, Steffi and Kou, Siqi and Su, Jiadi and Ma, Yan and Deng, Zhijie and Liu, Pengfei},
  journal={arXiv preprint arXiv:2505.22525},
  year={2025}
}

@inproceedings{visual_programming,
  title={Visual programming: Compositional visual reasoning without training},
  author={Gupta, Tanmay and Kembhavi, Aniruddha},
  booktitle={Proceedings of the IEEE/CVF conference on computer vision and pattern recognition},
  pages={14953--14962},
  year={2023}
}

@inproceedings{vipergpt,
  title={Vipergpt: Visual inference via python execution for reasoning},
  author={Sur{\'\i}s, D{\'\i}dac and Menon, Sachit and Vondrick, Carl},
  booktitle={Proceedings of the IEEE/CVF international conference on computer vision},
  pages={11888--11898},
  year={2023}
}

@misc{latent_reasoning_survey,
      title={A Survey on Latent Reasoning}, 
      author={Rui-Jie Zhu and Tianhao Peng and Tianhao Cheng and Xingwei Qu and Jinfa Huang and Dawei Zhu and Hao Wang and Kaiwen Xue and Xuanliang Zhang and Yong Shan and Tianle Cai and Taylor Kergan and Assel Kembay and Andrew Smith and Chenghua Lin and Binh Nguyen and Yuqi Pan and Yuhong Chou and Zefan Cai and Zhenhe Wu and Yongchi Zhao and Tianyu Liu and Jian Yang and Wangchunshu Zhou and Chujie Zheng and Chongxuan Li and Yuyin Zhou and Zhoujun Li and Zhaoxiang Zhang and Jiaheng Liu and Ge Zhang and Wenhao Huang and Jason Eshraghian},
      year={2025},
      eprint={2507.06203},
      archivePrefix={arXiv},
      primaryClass={cs.CL},
}

@article{coconut,
  title={Training large language models to reason in a continuous latent space},
  author={Hao, Shibo and Sukhbaatar, Sainbayar and Su, DiJia and Li, Xian and Hu, Zhiting and Weston, Jason and Tian, Yuandong},
  journal={arXiv preprint arXiv:2412.06769},
  year={2024}
}

@inproceedings{soft_thinking,
title={Soft Thinking: Unlocking the Reasoning Potential of {LLM}s in Continuous Concept Space},
author={Zhen Zhang and Xuehai He and Weixiang Yan and Ao Shen and Chenyang Zhao and Xin Eric Wang},
booktitle={The Thirty-ninth Annual Conference on Neural Information Processing Systems},
year={2026},
}

@inproceedings{pause_tokens,
title={Think before you speak: Training Language Models With Pause Tokens},
author={Sachin Goyal and Ziwei Ji and Ankit Singh Rawat and Aditya Krishna Menon and Sanjiv Kumar and Vaishnavh Nagarajan},
booktitle={The Twelfth International Conference on Learning Representations},
year={2024},
}

@inproceedings{quietstar,
title={Quiet-{ST}aR: Language Models Can Teach Themselves to Think Before Speaking},
author={Eric Zelikman and Georges Raif Harik and Yijia Shao and Varuna Jayasiri and Nick Haber and Noah Goodman},
booktitle={First Conference on Language Modeling},
year={2024},
}

@article{spot,
  title={SPOT: Span-level Pause-of-Thought for Efficient and Interpretable Latent Reasoning in Large Language Models},
  author={Chu, Yunlong and Shao, Minglai and Liu, Yuhang and Hao, Bing and Lin, Yumeng and Wang, Jialu and Wang, Ruijie},
  journal={arXiv preprint arXiv:2603.06222},
  year={2026}
}

@article{mcout,
  title={Multimodal chain of continuous thought for latent-space reasoning in vision-language models},
  author={Pham, Tan-Hanh and Ngo, Chris},
  journal={arXiv preprint arXiv:2508.12587},
  year={2025}
}

@inproceedings{lvr,
title={Latent Visual Reasoning},
author={Bangzheng Li and Ximeng Sun and Jiang Liu and Ze Wang and Jialian Wu and Xiaodong Yu and Emad Barsoum and Muhao Chen and Zicheng Liu},
booktitle={The Fourteenth International Conference on Learning Representations},
year={2026},
}

@article{ivtlr,
  title={Reasoning in the dark: Interleaved vision-text reasoning in latent space},
  author={Chen, Chao and Ma, Zhixin and Li, Yongqi and Hu, Yupeng and Wei, Yinwei and Li, Wenjie and Nie, Liqiang},
  journal={arXiv preprint arXiv:2510.12603},
  year={2025}
}

@article{covt,
  title={Chain-of-Visual-Thought: Teaching VLMs to See and Think Better with Continuous Visual Tokens},
  author={Qin, Yiming and Wei, Bomin and Ge, Jiaxin and Kallidromitis, Konstantinos and Fu, Stephanie and Darrell, Trevor and Wang, Xudong},
  journal={arXiv preprint arXiv:2511.19418},
  year={2025}
}

@article{latent_sketchpad,
title={Latent Sketchpad: Sketching Visual Thoughts to Elicit Multimodal Reasoning in MLLMs},
 author={Zhang, Huanyu and Wu, Wenshan and Li, Chengzu and Shang, Ning and Xia, Yan and Huang, Yangyu and Zhang, Yifan and Dong, Li and Zhang, Zhang and Wang, Liang and Tan, Tieniu and Wei, Furu},
journal={arXiv preprint arXiv:2510.24514},
year={2025}
}

@article{malt,
  title={Mull-Tokens: Modality-Agnostic Latent Thinking},
  author={Ray, Arijit and Abdelkader, Ahmed and Mao, Chengzhi and Plummer, Bryan A and Saenko, Kate and Krishna, Ranjay and Guibas, Leonidas and Chu, Wen-Sheng},
  journal={arXiv preprint arXiv:2512.10941},
  year={2025}
}

@article{reasoning_mind,
  title={Reasoning within the mind: Dynamic multimodal interleaving in latent space},
  author={Liu, Chengzhi and Yang, Yuzhe and Fan, Yue and Wei, Qingyue and Liu, Sheng and Wang, Xin Eric},
  journal={arXiv preprint arXiv:2512.12623},
  year={2025}
}

@inproceedings{tte,
title={Think Then Embed: Generative Context Improves Multimodal Embedding},
author={Xuanming Cui and Jianpeng Cheng and Hong-You Chen and Satya Narayan Shukla and Abhijeet Awasthi and Xichen Pan and Chaitanya Ahuja and Shlok Kumar Mishra and Taipeng Tian and Qi Guo and Ser-Nam Lim and Aashu Singh and Xiangjun Fan},
booktitle={The Fourteenth International Conference on Learning Representations},
year={2026},
url={https://openreview.net/forum?id=AKXXMK5YTI}
}

@article{hive,
  title={Multimodal Latent Reasoning via Hierarchical Visual Cues Injection},
  author={Zhang, Yiming and Yan, Qiangyu and Jiang, Borui and Han, Kai},
  journal={arXiv preprint arXiv:2602.05359},
  year={2026}
}

@article{lavit,
  title={LaViT: Aligning Latent Visual Thoughts for Multi-modal Reasoning},
  author={Wu, Linquan and Jiang, Tianxiang and Dong, Yifei and Yang, Haoyu and Zhang, Fengji and Meng, Shichaang and Xuan, Ai and Song, Linqi and Keung, Jacky},
  journal={arXiv preprint arXiv:2601.10129},
  year={2026}
}

@article{continuous_visual_thought_review,
  title={Thinking with images for multimodal reasoning: Foundations, methods, and future frontiers},
  author={Su, Zhaochen and Xia, Peng and Guo, Hangyu and Liu, Zhenhua and Ma, Yan and Qu, Xiaoye and Liu, Jiaqi and Li, Yanshu and Zeng, Kaide and Yang, Zhengyuan and others},
  journal={arXiv preprint arXiv:2506.23918},
  year={2025}
}

@inproceedings{monet,
      title={Monet: Reasoning in Latent Visual Space Beyond Images and Language}, 
      author={Qixun Wang and Yang Shi and Yifei Wang and Yuanxing Zhang and Pengfei Wan and Kun Gai and Xianghua Ying and Yisen Wang},
      year={2026},
      booktitle={CVPR}
}

@inproceedings{v_star,
  title={V?: Guided visual search as a core mechanism in multimodal llms},
  author={Wu, Penghao and Xie, Saining},
  booktitle={Proceedings of the IEEE/CVF Conference on Computer Vision and Pattern Recognition},
  pages={13084--13094},
  year={2024}
}

@inproceedings{blink,
  title={Blink: Multimodal large language models can see but not perceive},
  author={Fu, Xingyu and Hu, Yushi and Li, Bangzheng and Feng, Yu and Wang, Haoyu and Lin, Xudong and Roth, Dan and Smith, Noah A and Ma, Wei-Chiu and Krishna, Ranjay},
  booktitle={European Conference on Computer Vision},
  pages={148--166},
  year={2024},
  organization={Springer}
}

@inproceedings{mmvp,
  title={Eyes wide shut? exploring the visual shortcomings of multimodal llms},
  author={Tong, Shengbang and Liu, Zhuang and Zhai, Yuexiang and Ma, Yi and LeCun, Yann and Xie, Saining},
  booktitle={Proceedings of the IEEE/CVF conference on computer vision and pattern recognition},
  pages={9568--9578},
  year={2024}
}

@inproceedings{math,
  author    = {Lu, Pan and Bansal, Hritik and Xia, Tony and Liu, Jiacheng and Li, Chunyuan and Hajishirzi, Hannaneh and Cheng, Hao and Chang, Kai-Wei and Galley, Michel and Gao, Jianfeng},
  title     = {MathVista: Evaluating Mathematical Reasoning of Foundation Models in Visual Contexts},
  booktitle={International Conference on Learning Representations (ICLR)},
  year      = {2024}
}

@inproceedings{chartqa,
  title={Chartqa: A benchmark for question answering about charts with visual and logical reasoning},
  author={Masry, Ahmed and Do, Xuan Long and Tan, Jia Qing and Joty, Shafiq and Hoque, Enamul},
  booktitle={Findings of the association for computational linguistics: ACL 2022},
  pages={2263--2279},
  year={2022}
}

@misc{openai2024gpt4o,
  title        = {Hello GPT-4o},
  author       = {{OpenAI}},
  year         = {2024},
  howpublished = {\url{https://openai.com/index/hello-gpt-4o/}},
  note         = {Accessed: 2026-05-03}
}

@article{gemini2024gemini15,
  title   = {Gemini 1.5: Unlocking multimodal understanding across millions of tokens of context},
  author  = {{Gemini Team}},
  journal = {arXiv preprint arXiv:2403.05530},
  year    = {2024}
}

@article{bai2025qwen25vl,
  title   = {Qwen2.5-VL Technical Report},
  author  = {Bai, Shuai and Chen, Keqin and Liu, Xuejing and others},
  journal = {arXiv preprint arXiv:2502.13923},
  year    = {2025}
}

@article{hong2025glm45v,
  title   = {GLM-4.5V and GLM-4.1V-Thinking: Towards Versatile Multimodal Understanding and Reasoning},
  author  = {Hong, Wenyi and others},
  journal = {arXiv preprint arXiv:2507.01006},
  year    = {2025}
}

@article{feng2025visually,
  title={Visually Prompted Benchmarks Are Surprisingly Fragile},
  author={Feng, Haiwen and Lian, Long and Dunlap, Lisa and Shu, Jiahao and Wang, XuDong and Wang, Renhao and Darrell, Trevor and Suhr, Alane and Kanazawa, Angjoo},
  journal={arXiv preprint arXiv:2512.17875},
  year={2025}
}

@inproceedings{mmmu,
  title={Mmmu: A massive multi-discipline multimodal understanding and reasoning benchmark for expert agi},
  author={Yue, Xiang and Ni, Yuansheng and Zhang, Kai and Zheng, Tianyu and Liu, Ruoqi and Zhang, Ge and Stevens, Samuel and Jiang, Dongfu and Ren, Weiming and Sun, Yuxuan and others},
  booktitle={Proceedings of the IEEE/CVF conference on computer vision and pattern recognition},
  pages={9556--9567},
  year={2024}
}

@article{lacot,
  title={Latent chain-of-thought for visual reasoning},
  author={Sun, Guohao and Hua, Hang and Wang, Jian and Luo, Jiebo and Dianat, Sohail and Rabbani, Majid and Rao, Raghuveer and Tao, Zhiqiang},
  journal={arXiv preprint arXiv:2510.23925},
  year={2025}
}
\bibliographystyle{plainnat}

\appendix
\newpage
\section{Technical appendices and supplementary material}
\subsection{Diagnostic Evaluation under Misleading Captions}
\label{app:misleading_caption}

To further verify whether SLVR relies on visual evidence rather than textual shortcuts, we construct a diagnostic evaluation set with misleading textual contexts.
We sample 300 image-text examples from COVT~\cite{covt} that do not overlap with our training set.
For each example, we use an off-the-shelf VLM-assisted rewriting procedure to identify the answer-critical visual fact in the textual context and deliberately replace it with an incorrect but plausible alternative.
The image and question are kept unchanged, so the correct answer is still determined by the image rather than the modified caption.
For instance, if the image contains four green apples, the misleading context may state that there are two green apples.
This creates a controlled conflict between the textual context and the visual evidence.

We evaluate four variants: the Qwen2.5-VL-7B base model, the VDI-only model, LVR~\cite{lvr}, and the full SLVR model.
This comparison tests whether each model can recover the correct answer from the image when the caption provides a plausible but incorrect shortcut.
We report two metrics.
\textit{Misleading accuracy} measures the exact-match accuracy under the perturbed textual context.
\textit{Caption-following error} measures the percentage of examples where the model predicts the incorrect answer implied by the misleading caption.
A lower caption-following error indicates stronger resistance to textual shortcuts.

\begin{table}[t]
\centering
\small
\setlength{\tabcolsep}{5pt}
\renewcommand{\arraystretch}{1.0}
\caption{
\textbf{Diagnostic evaluation under misleading captions.}
We evaluate 300 non-overlapping COVT examples where the answer-critical visual fact in the textual context is deliberately replaced with an incorrect but plausible alternative.
The image and question remain unchanged.
A grounded model should recover the correct answer from the image rather than follow the misleading caption.
}
\begin{tabular}{lccc}
\toprule
Method 
& Clean Acc. $\uparrow$ 
& Misleading Acc. $\uparrow$ 
& Caption-following Err. $\downarrow$ \\
\midrule
Qwen2.5-VL-7B 
& 72.3 
& 41.7 
& 46.0 \\
VDI-only 
& 78.0 
& 58.3 
& 27.7 \\
LVR~\cite{lvr}
& 80.7 
& 62.0 
& 23.3 \\
SLVR 
& \textbf{84.3} 
& \textbf{70.7} 
& \textbf{14.7} \\
\bottomrule
\end{tabular}
\label{tab:misleading_caption_appendix}
\end{table}

\subsection{Hyperparameter Analysis}
\label{app:hyperparameter_analysis}

Table~\ref{tab:hyperparameter_analysis} analyzes two key hyperparameter choices.
For evidence selection, setting $\tau_{\mathrm{evi}}=0$ keeps nearly all candidate patches and introduces noisy visual evidence, while a high threshold may discard useful patches and more frequently trigger the top-ranked fallback.
We use $\tau_{\mathrm{evi}}=0.5$ and $K_{\min}=5$ in all experiments, which provides the best balance between evidence recall and precision.

For auxiliary loss weighting, we weight each objective according to its role in the reasoning pipeline.
We down-weight the plan loss since high-level reasoning plans are semantically flexible and should not over-constrain the model.
Grounding and patch losses are kept at standard weight because they provide concrete visual supervision.
The answer embedding loss is weighted moderately to align latent states with textual answers without dominating the language modeling objective.
In contrast, we up-weight the answer-opening and short-answer losses, since they respectively control the transition from latent reasoning to the explicit \texttt{<answer>} segment and the correctness of the final prediction.
The lower performance of uniform weighting and removing these tailored weights shows that such role-aware balancing is important for SLVR.

\begin{table*}[t]
\centering
\small
\setlength{\tabcolsep}{6pt}
\renewcommand{\arraystretch}{1.08}
\caption{
\textbf{Hyperparameter analysis of SLVR.}
Left: sensitivity to the evidence threshold $\tau_{\mathrm{evi}}$ for patch selection.
When $\tau_{\mathrm{evi}}=0$, the selection degenerates to using nearly all candidate patches, introducing noisy visual evidence.
If fewer than $K_{\min}$ patches satisfy the threshold, we fall back to the top-ranked candidates.
Right: ablation on auxiliary loss weighting strategies.
All variants are evaluated on \textbf{MMVP} and \textbf{V* Overall}.
}
\label{tab:hyperparameter_analysis}
\resizebox{\textwidth}{!}{
\begin{tabular}{lcc|lcc}
\toprule
\multicolumn{3}{c|}{\textbf{Evidence Threshold $\tau_{\mathrm{evi}}$}}
& \multicolumn{3}{c}{\textbf{Auxiliary Loss Weighting}} \\
\midrule
\textbf{Threshold} 
& \textbf{MMVP} 
& \textbf{V* Overall}
& \textbf{Weight Setting}
& \textbf{MMVP}
& \textbf{V* Overall} \\
\midrule
0.0
& 73.6 & 83.9
& Uniform weights
& 74.8 & 84.3 \\

0.3
& 75.0 & 84.4
& w/o up-weighted answer
& 75.2 & 84.6 \\

0.5
& \cellcolor{green!8}\textbf{76.1}
& \cellcolor{green!8}\textbf{85.2}
& w/o down-weighted plan loss
& 75.0 & 84.5 \\

0.7
& 75.4 & 84.6
& \cellcolor{green!8}\textbf{Ours}
& \cellcolor{green!8}\textbf{76.1}
& \cellcolor{green!8}\textbf{85.2} \\
\bottomrule
\end{tabular}
}
\end{table*}

\subsection{Comparison with Latent Multimodal Reasoning Methods}
\label{app:latent_comparison}

Table~\ref{tab:latent_comparison} summarizes how SLVR differs from representative latent multimodal reasoning methods.
We emphasize that several prior methods introduce meaningful forms of structure, including continuous thought tokens, reconstructed visual tokens, interleaved vision-text states, latent visual embeddings, and visual scratchpads.
Our comparison focuses on whether a method explicitly decomposes latent reasoning into the same functional stages used by SLVR: planning, grounding, evidence selection, and reasoning integration.

\begin{table*}[t]
\centering
\small
\setlength{\tabcolsep}{5pt}
\renewcommand{\arraystretch}{1.08}
\caption{
\textbf{High-level comparison with representative latent multimodal reasoning methods.}
Abbreviations: \textbf{Latent Sup.} = continuous latent states with explicit latent supervision;
\textbf{Visual Feat.} = explicit use, selection, injection, or reconstruction of visual features during latent reasoning;
\textbf{Grounding Sup.} = explicit region-level grounding supervision, such as boxes or ROI targets;
\textbf{Typed} = typed plan/grounding/evidence/integration stages;
\textbf{Ans. Evi. Sup.} = answer-specific evidence supervision.
The columns distinguish design choices rather than overall capability.
}
\resizebox{\textwidth}{!}{
\begin{tabular}{lccccc}
\toprule
\textbf{Method}
& \textbf{Latent Sup.}
& \textbf{Visual Feat.}
& \textbf{Grounding Sup.}
& \textbf{Typed}
& \textbf{Ans. Evi. Sup.} \\
\midrule

Multimodal Coconut~\cite{mcout}
& \checkmark
& --
& --
& --
& -- \\

LVR~\cite{lvr}
& \checkmark
& \checkmark
& \checkmark
& --
& \checkmark \\

IVT-LR~\cite{ivtlr}
& \checkmark
& \checkmark
& --
& --
& -- \\

LaCoT~\cite{lacot}
& \checkmark
& \checkmark
& --
& --
& -- \\

COVT~\cite{covt}
& \checkmark
& \checkmark
& \checkmark
& --
& \checkmark \\

Latent Sketchpad~\cite{latent_sketchpad}
& \checkmark
& \checkmark
& --
& --
& -- \\

MoNet~\cite{monet}
& \checkmark
& \checkmark
& --
& --
& -- \\

LaViT~\cite{lavit}
& \checkmark
& \checkmark
& --
& --
& -- \\

SLVR (Ours)
& \checkmark
& \checkmark
& \checkmark
& \checkmark
& \checkmark \\
\bottomrule
\end{tabular}
}
\label{tab:latent_comparison}
\end{table*}

\subsection{Details of Visual Dependency Induction}
\label{app:vdi}

This appendix shows the data format used in the Visual Dependency Induction stage.
The purpose of this stage is to create image-conditioned multiple-choice examples where the rewritten text remains aligned with the image but does not directly contain the answer.
This forces the model to recover the missing visual fact from the image rather than relying on textual shortcuts or answer priors.

\begin{figure}[t]
    \centering
    \includegraphics[width=\linewidth]{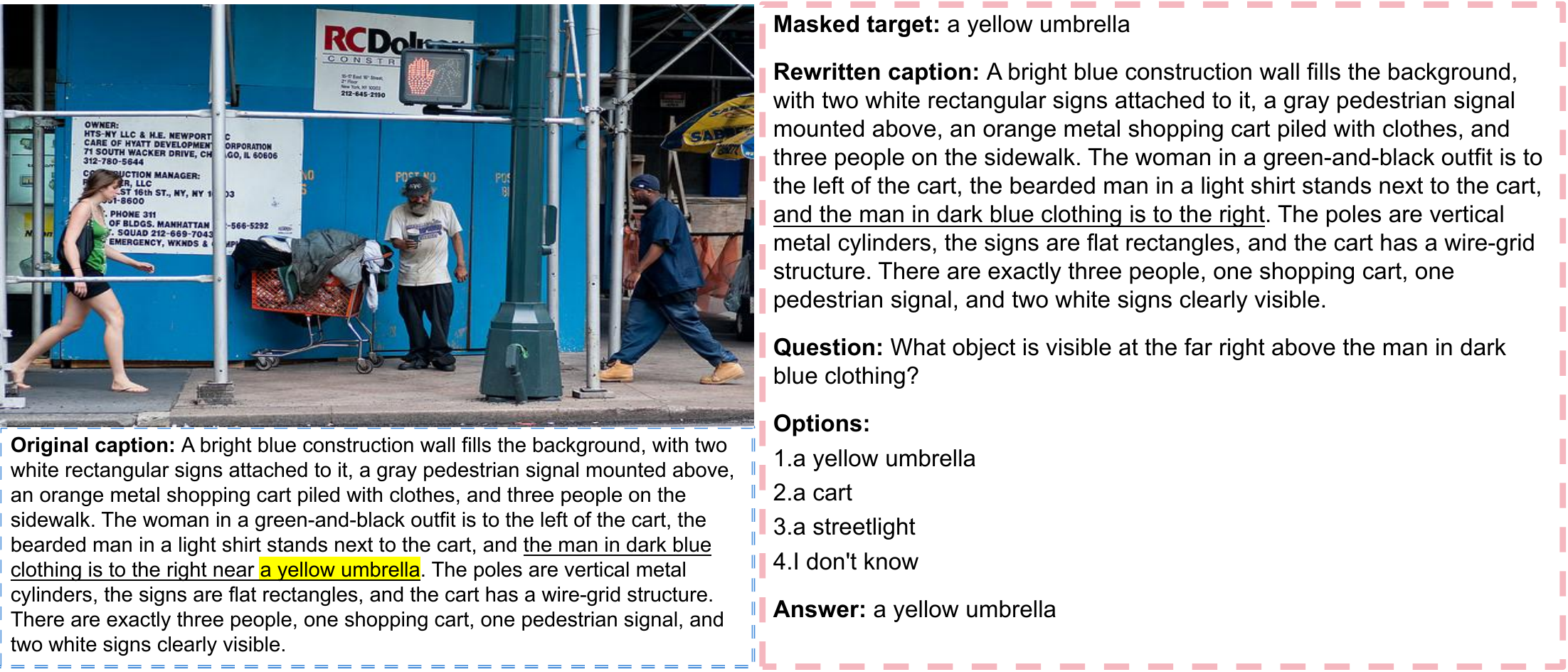}
    \caption{
    \textbf{Example training instance for Visual Dependency Induction.}
    Each instance contains an original caption, a masked target, a rewritten caption, a question, answer options, and the correct answer.
    The target visual fact, \textit{a yellow umbrella}, is removed from the rewritten caption while the remaining image-caption context is preserved.
    The model must ground the question in the image to recover the missing fact.
    }
    \label{fig:vdi_example}
\end{figure}

\paragraph{Visual Dependency Induction data format.}
Each training instance in the Visual Dependency Induction stage is organized as an image-conditioned multiple-choice example.
As shown in Figure~\ref{fig:vdi_example}, each example contains an image, an original caption, a masked target, a rewritten caption, a question, a set of answer options, and the correct answer.
The original caption provides a complete image-grounded description and may explicitly mention the target visual fact.
The masked target specifies the visual evidence to be removed from the text.
The rewritten caption preserves the remaining image-caption context but removes this target fact, so that the answer is no longer available from text alone.
The question is then written to query the removed target fact, and the options contain one correct answer together with visually plausible distractors.
Therefore, during training, the model must condition on both the rewritten caption and the image to recover the correct answer.

\paragraph{Example.}
Figure~\ref{fig:vdi_example} shows one Visual Dependency Induction training instance.
The original caption explicitly mentions the target object, ``a yellow umbrella.''
The rewritten caption removes this target while preserving the other scene information, including the blue construction wall, white signs, pedestrian signal, shopping cart, and people on the sidewalk.
The resulting question asks the model to identify the object visible at the far right above the man in dark blue clothing.
Since the rewritten caption no longer contains the target object, the answer can only be recovered by looking at the image.

\paragraph{Instance fields.}
The example in Figure~\ref{fig:vdi_example} contains the following fields.

\paragraph{Original caption.}
A bright blue construction wall fills the background, with two white rectangular signs attached to it, a gray pedestrian signal mounted above, an orange metal shopping cart piled with clothes, and three people on the sidewalk.
The woman in a green-and-black outfit is to the left of the cart, the bearded man in a light shirt stands next to the cart, and the man in dark blue clothing is to the right near a yellow umbrella.
The poles are vertical metal cylinders, the signs are flat rectangles, and the cart has a wire-grid structure.
There are exactly three people, one shopping cart, one pedestrian signal, and two white signs clearly visible.

\paragraph{Masked target.}
A yellow umbrella.

\paragraph{Rewritten caption.}
A bright blue construction wall fills the background, with two white rectangular signs attached to it, a gray pedestrian signal mounted above, an orange metal shopping cart piled with clothes, and three people on the sidewalk.
The woman in a green-and-black outfit is to the left of the cart, the bearded man in a light shirt stands next to the cart, and the man in dark blue clothing is to the right.
The poles are vertical metal cylinders, the signs are flat rectangles, and the cart has a wire-grid structure.
There are exactly three people, one shopping cart, one pedestrian signal, and two white signs clearly visible.

\paragraph{Question.}
What object is visible at the far right above the man in dark blue clothing?

\paragraph{Options.}
\begin{enumerate}
    \item A yellow umbrella.
    \item A cart.
    \item A streetlight.
    \item I don't know.
\end{enumerate}

\paragraph{Answer.}
A yellow umbrella.

\paragraph{Data construction pipeline.}
We construct the Visual Dependency Induction data with a teacher-model-assisted generation and filtering pipeline.
The teacher model is stronger than the student model used for training, and is used only to generate supervision data.
Given an image-caption pair, the teacher model first identifies a concrete visual target fact that is directly observable in the image and is also mentioned in the caption.
The target fact is selected from common visually grounded categories, including object presence, color, counting, spatial relations, and physical or structural attributes.
We require the target fact to be visually verifiable, unambiguous, and difficult to infer from the remaining text alone.

After selecting the target fact, the teacher model rewrites the caption by removing only the target information while preserving the remaining image-caption context.
The rewritten caption is required to remain fluent and aligned with the image, but it must not reveal the answer through synonyms, paraphrases, indirect descriptions, or commonsense shortcuts.
The teacher model then generates a multiple-choice question that queries the removed visual fact.
Each question contains one correct answer, visually plausible hard distractors, and an ``I don't know'' option.
The distractors are chosen to be plausible under the same image context but incorrect with respect to the queried visual evidence, so that the model cannot answer reliably from the rewritten caption alone.

\paragraph{Automatic verification.}
To reduce caption leakage and ambiguous supervision, we use a second VLM/LLM-based verifier to review the generated samples.
The verifier checks whether: 
(1) the rewritten caption removes the target fact;
(2) the rewritten caption does not leak the answer through synonyms or indirect clues;
(3) the question is answerable from the image;
(4) the correct answer is unique among the options;
(5) the distractors are plausible but visually incorrect; and
(6) the answer cannot be inferred from the rewritten caption alone.
Samples that fail any of these checks are discarded.
In our implementation, we generate the initial VDI candidates with Gemini and use GPT-based verification to filter invalid examples.
Starting from 46.5K generated candidates, the verification stage retains approximately 20K high-quality examples.

\paragraph{Human inspection.}
In addition to automatic verification, we manually inspect a subset of the generated data to estimate data quality.
We review approximately 1K examples sampled from the generated pool, focusing on caption leakage, answer uniqueness, visual ambiguity, and whether the question truly requires image grounding.
This inspection confirms that the retained examples generally follow the intended VDI format: the rewritten caption preserves the scene context while removing the answer-critical visual fact, and the correct answer is uniquely recoverable from the image.

\paragraph{Category balance.}
We balance the final VDI data across major visual-dependency categories to avoid overfitting to a single type of visual cue.
The final data approximately covers object presence, color, counting, spatial relations, and physical or structural attributes.
Each category contributes roughly a comparable portion of the training set.
This balanced design encourages the model to learn different forms of image-text dependency rather than relying on one dominant shortcut type.

\subsection{Visual-CoT Data Format and Latent Supervision}
\label{app:visual_cot_format}

We use Visual-CoT for supervised training. Each example is stored as one JSONL row and contains an image-grounded question-answer pair, a grounding bounding box, and textual reasoning annotations.

Each data row contains the following fields:
\begin{itemize}
    \item \texttt{question}: the natural-language visual question.
    \item \texttt{answer}: the short answer used as the final explicit answer target.
    \item \texttt{full\_answer}: a complete-sentence answer, used as latent answer supervision.
    \item \texttt{image}: the image filename.
    \item \texttt{width}, \texttt{height}: original image size.
    \item \texttt{bboxs}: a list of relevant visual bounding boxes in pixel coordinates, formatted as \([x_0,y_0,x_1,y_1]\). We use the first box as the grounding target.
    \item \texttt{reasoning}: symbolic GQA-style reasoning operations, such as \texttt{select}, \texttt{relate}, \texttt{query}, and attribute filters.
    \item \texttt{thought}: a natural-language chain-of-thought style explanation describing the visual reasoning process.
    \item \texttt{dataset} and \texttt{split}: metadata indicating the source dataset and split.
\end{itemize}

An example row is:

\begin{Verbatim}[breaklines=true,breakanywhere=true,fontsize=\small]
{
  "question": "Who is wearing a shirt?",
  "answer": "girl",
  "full_answer": "The girl is wearing a shirt.",
  "image": "2331819.jpg",
  "width": 500,
  "height": 375,
  "bboxs": [[214, 0, 433, 374]],
  "dataset": "gqa",
  "split": "train",
  "plan": "1. We are looking for someone who is wearing a shirt in the image.\n2. First, let's identify all the people in the image.\n3. Next, we need to determine who among the people is wearing a shirt.\n4. After identifying the person, we can state the answer based on the observation from the image."
}
\end{Verbatim}
\paragraph{Latent-state construction.}
Instead of exposing the full reasoning text as ordinary output tokens, we convert different supervision signals into typed latent blocks. The assistant-side training sequence is constructed as:
\[
\texttt{<plan\_lvr\_start>}~\texttt{<lvr>}^{K_1}~\texttt{<plan\_lvr\_end>}
\]
\[
\texttt{<bbox\_lvr\_start>}~\texttt{<lvr>}^3~\texttt{<bbox\_lvr\_end>}
\]
\[
\texttt{<patch\_lvr\_start>}~\texttt{<lvr>}^{K_3}~\texttt{<patch\_lvr\_end>}
\]
\[
\texttt{<full\_lvr\_start>}~\texttt{<lvr>}^{K_4}~\texttt{<full\_lvr\_end>}
\]
\[
\texttt{<answer>}~a_{\mathrm{short}}~\texttt{</answer>}.
\]

The four latent blocks correspond to different supervision sources:

\begin{itemize}
    \item \textbf{Plan latent block.} The \texttt{thought} field is tokenized and truncated to at most \(K_1\) tokens. These token embeddings are used as latent targets for the plan block.
    \item \textbf{Bounding-box latent block.} The selected bounding box is normalized by image width and height, converted from \((x_0,y_0,x_1,y_1)\) to \((c_x,c_y,w,h)\), and represented by three latent targets: center coordinate \((c_x,c_y)\), width \(w\), and height \(h\).
    \item \textbf{Patch latent block.} The normalized box is mapped to Qwen2.5-VL image patch indices. Candidate patch embeddings inside the box are ranked by cosine similarity to the mean embedding of the full answer. The top patches are used as latent visual targets.
    \item \textbf{Full-answer latent block.} The \texttt{full\_answer} field is tokenized and used as latent answer-level supervision.
\end{itemize}


\newpage

\end{document}